\pdfoutput=1

\documentclass[11pt]{article}

\usepackage[]{LAW2021}

\usepackage{times}
\usepackage{latexsym}

\usepackage[T1]{fontenc}

\usepackage[utf8]{inputenc}

\usepackage{microtype}

\usepackage{booktabs}
\usepackage{graphicx}
\usepackage{xspace}
\usepackage{fdsymbol}
\usepackage{multirow}

\newcommand{\sesame}{\textsc{open-sesame}\xspace}

\newcommand{\draftonly}[1]{#1}
\renewcommand{\draftonly}[1]{}    

\title{Sister Help: Data Augmentation for Frame-Semantic Role Labeling}

\author{
Ayush Pancholy$^\vardiamondsuit\;\;\;$
Miriam R. L. Petruck$^\varheartsuit\;\;\;$ 
Swabha Swayamdipta$^\clubsuit\;\;\;$ \\
$^\vardiamondsuit$ University of California, Berkeley \\
$^\varheartsuit$ International Computer Science Institute, Berkeley \\
$^\clubsuit$ Allen Institute for Artificial Intelligence \\
\texttt{\{ayush.pancholy@,miriamp@icsi.\}berkeley.edu}\\
\texttt{swabhas@allenai.org}
}

\begin{document}
\maketitle

\begin{abstract}

While FrameNet is widely regarded as a rich resource of semantics in natural language processing, a major criticism concerns its lack of coverage and the relative paucity of its labeled data compared to other commonly used lexical resources such as PropBank and VerbNet. This paper reports on a pilot study to address these gaps. We propose a data augmentation approach, which uses existing frame-specific annotation to automatically annotate other lexical units of the same frame which are unannotated. Our rule-based approach defines the notion of a \textbf{sister lexical unit} and generates frame-specific augmented data for training. We present experiments on frame-semantic role labeling which demonstrate the importance of this data augmentation: we obtain a large improvement to prior results on frame identification and argument identification for FrameNet, utilizing both full-text and lexicographic annotations under FrameNet. Our findings on data augmentation highlight the value of automatic resource creation for improved models in frame-semantic parsing.
\end{abstract}
\section{Introduction}
\label{sec:intro}

Among the challenges to Natural Language Processing (NLP) systems is access to sufficient and accurate information about the mapping between form and meaning in language \citep{benderkoller20climbing}. 
While the number of resources that provide such information has increased in recent decades, producing these resources remains labor-intensive and costly.
This is particularly true for FrameNet (\textbf{FN}; \citealp{Ruppenhofer:16}), a rich resource of semantic annotations, popularized for semantic role labeling (\textbf{SRL}) since the pioneering work of \citet{Dans02}.

Nevertheless, FrameNet still relies (almost exclusively) on expert manual linguistic annotation, resulting in far fewer annotations than other resources that are able to scale up the annotation process easily \cite{fitzgerald-etal-2018-large}.
Moreover, the fine-grained annotations in FrameNet make them harder to produce compared to other lexical resources such as PropBank \citep{propbank} and VerbNet \citep{verbnet}.
As a result, while SRL has continued to be a mainstay of core NLP, with applications in event extraction, participant tracking, machine translation, and question-answering, far fewer SRL systems use FrameNet for training, in contrast to PropBank.\footnote{PropBank annotations were also included in CoNLL shared tasks in 2005 \cite{carreras-marquez-2005-introduction} and 2012 \cite{pradhan-etal-2012-conll}, and in OntoNotes \cite{pradhan-etal-2013-towards}.}

\begin{figure}[t!]
\centering
\includegraphics[width=\linewidth]{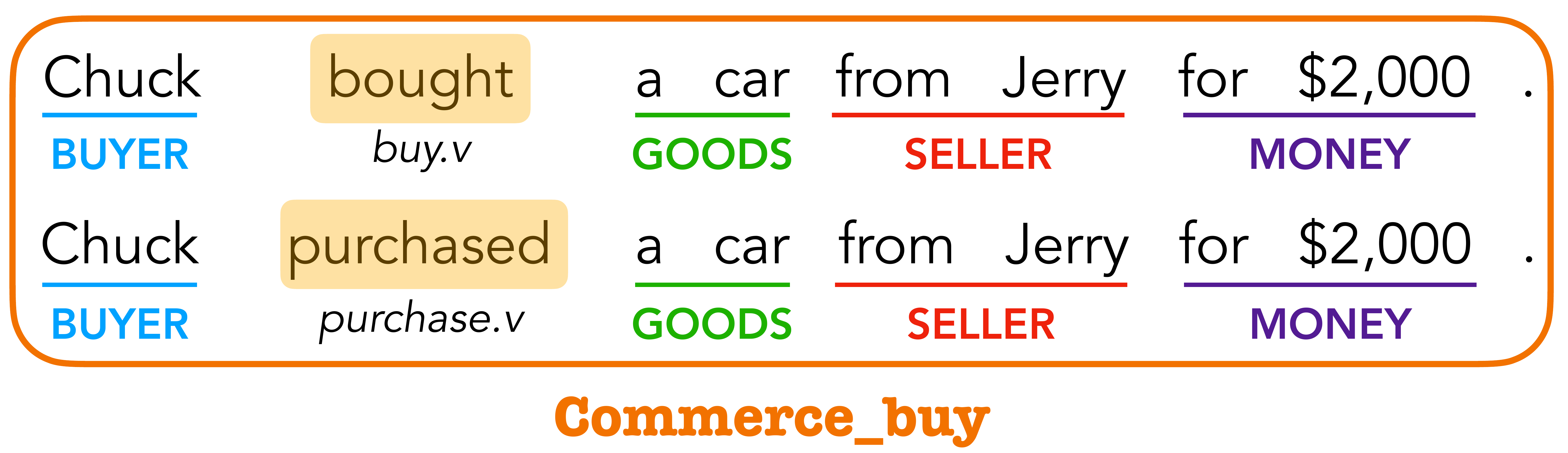}
\caption{Illustration of Sister Lexical Units (\textbf{Sister LUs}) with FrameNet annotation for the verbs \textit{buy} and \textit{purchase}, which are defined in terms of the \texttt{Commerce\_buy} frame.
The arguments of both LUs are identical, both realizing the frame elements (FEs) \textsc{{\textcolor{blue}{Buyer}}}, \textsc{{\textcolor{green}{Goods}}}, \textsc{{\textcolor{red}{Seller}}}, and \textsc{{\textcolor{violet}{Money}}}.
Targets in the sentence that trigger the frame are the words \textit{bought} and \textit{purchased}, respectively.
}
\label{fig:illustration}
\end{figure}

Our work seeks to address this low resource problem in FrameNet in English by automatic data augmentation.
We leverage the FrameNet hierarchy where multiple lexical units are associated with the same frame, but might have a different number of annotations associated with them (\S\ref{sec:background}).
We propose a rule-based approach to transfer annotations among sister LUs; see Fig. \ref{fig:illustration} for an illustration. 
We implement and extend the proposal by 
\citet{FSautoanno19} to generate new annotation for previously unannotated LUs in the same frame, by exploiting the already identified target and frame of a given lemma.
We hypothesize that our rule-based approach to transfer \textit{{frame-specific}} annotation to generate semantic role labels for example sentences of lexical units (\textbf{LU}s) for which none exist, can indeed result in a FrameNet with higher coverage annotations, a state of affairs that 
can help in training frame-semantic role labeling systems (\S\ref{sec:methodology}).
Our experiments show that this is indeed the case; models trained on augmented data are able to surpass a strong baseline (\sesame; \citealp{DBLP:journals/corr/SwayamdiptaTDS17}) for frame and argument identification (\S\ref{sec:experiments}).
We also provide a summary of the types of linguistic errors in the augmented data and the SRL task.

While our work provides a proof of concept, its novelty lies in exploiting frame-specific annotation to produce augmented data yielding new annotation.
Our implementation and augmented data are publicly available.\footnote{\url{https://github.com/ayush-pancholy/sister-help}}



\section{Background}
\label{sec:background}

\subsection{FrameNet}
FrameNet \citep{Ruppenhofer:16} is a research and resource development project in corpus-based computational lexicography grounded in the principles of \textbf{frame semantics} \citep{Fillmore:85}.
One of
the goals of this effort 
is documenting the \textbf{valences}, i.e., the syntactic and semantic combinatorial possibilities of each item analyzed.
These valence descriptions provide critical information on the mapping between form and meaning; NLP and more broadly, natural language understanding (NLU) require such mapping. 

At the heart of the work is the \textbf{semantic frame}, a script-like knowledge structure that facilitates inference within and across events, situations, relations, etc. 
FrameNet defines a semantic frame in terms of its \textbf{frame elements} (FEs), or participants in the scene that the frame captures; a \textbf{lexical unit} (LU)\label{LU} is a pairing of a lemma and a frame, thus characterizing that LU in terms of the frame that it evokes.
Valence descriptions derive from the annotation of FEs, i.e. \textbf{semantic roles}, on example sentences that illustrate the linguistic manifestation of the participants in a scene for the target of analysis.
The first sentence in Figure~\ref{fig:illustration} illustrates annotation with respect to the verb \textit{buy}, which FN defines in terms of the \texttt{Commerce\_buy} frame, whose FEs are a \textsc{{\textcolor{blue}{Buyer}}}, \textsc{{\textcolor{red}{Seller}}}, \textsc{{\textcolor{green}{Goods}}}, and \textsc{{\textcolor{violet}{Money}}}.\footnote{Frame names appear in \texttt{Typewriter} font; FE names are in \textsc{Small Caps}; and LUs are in \textit{italics}.}





The FrameNet\footnote{\url{https://framenet.icsi.berkeley.edu/fndrupal/framenet_request_data}} database (version 1.7) holds 1,224 frames, 10,478 frame-specific FEs (in lexical frames),\footnote{FrameNet distinguishes between lexical and non-lexical frames, where the latter do not have associated annotation sets.} over 13,500 LUs.
The primary annotations are in the form of nearly 203,000 manually annotated sentences providing information about the mapping between form and meaning in English; these are called the \textbf{lexicographic} annotations. 
A subset of the sentences with lexicographic annotations contain annotations for all possible frames of meaning in them; FrameNet calls this subset full-text annotations.
Despite FrameNet's costly annotation efforts, 38\% of the over 13,500 LUs in the database remain without annotation. 
That is, a significant amount of LUs in the database have no associated annotated sentences to serve as training data for downstream NLP applications.

\subsection{Frame Semantic Role Labeling}
\label{SRL}
Semantic role labeling (\textbf{SRL)} involves automatically labeling who did what to whom in a text document.
SRL systems facilitate developing {NLP} applications, like question answering
\citep{shenlap07using}, machine translation \citep{pedersen01},
and text summarization \citep{Han16}, to name but a few.
Frame semantic role labeling (\textbf{frame-SRL)} is a special type of SRL where the task is to identify target tokens (which evoke a frame of meaning), the frame itself (of target LUs) and FEs, or semantic roles, in text. 
The relationship between frames and their respective FEs requires that the system first identify the correct frame (FrameID) to identify the correct FEs (ArgID).
Identifying the frame incorrectly necessarily means incorrect FE identification.\footnote{\citet{semeval/Baker07} also considered correct identification of the spans of a dependent of a lexical unit, even if the frame and by definition the FEs were identified incorrectly.}
In practice, frame-SRL is usually implemented as a pipeline of structured prediction \cite{smith2011} tasks: identifying targets and LUs, frames, and finally the FEs \cite{das10probabilistic}.

Using an early version of FN data \citep{FNearly}, \citet{Dans02} developed the first SRL system, which also
initiated SRL as a now well-recognized task in the field.
Recent years have seen the development of several (relatively) high-performing SRL systems. 
SEMAFOR \citep{das2014frame}
uses a pipeline of discrete, manually designed feature-based linear classifiers for target identification, frame identification (FrameID), and argument identification (ArgID). 
The core constraints of frame-semantic analyses, e.g. not repeating core FEs, are satisfied via
integer linear programming.
PathLSTM \citep{Roth:16} uses neural features to embed path relationships between frames and FEs, in a pipeline similar to that of SEMAFOR.
\citet{yang-mitchell-2017-joint} use a joint model to leverage Propbank SRL annotations to improve frame-SRL.
\sesame \cite{DBLP:journals/corr/SwayamdiptaTDS17,swayamdiptaEtAl18syntactic} uses an unconstrained, neural approach in a pipeline 
like SEMAFOR and PathLSTM. 
Continuous representations and a sophisticated global model based on semi-Markov conditional random fields \citep{conf/nips/SarawagiC04} improve identifying arguments.
We employ \sesame\footnote{\url{https://github.com/swabhs/open-sesame}} as the baseline of choice in our experiments (\S\ref{sec:experiments}).
With the exception of a few approaches, like that of \citet{yang-mitchell-2017-joint}, most prior work uses only the full-text annotations for frame-SRL; we present experiments considering the larger set of lexicographic annotations as well.






\section{Method: Data Augmentation with Sister LUs}
\label{sec:methodology}


The relationship among LUs in a frame
motivates examining their use for paraphrasing, where one LU replaces another resulting in an alternative phrasing of a sentence.
\citet{mutaphrase07} produced paraphrases based on the relationship among LUs of a frame or closely related frames, augmenting a meager data set for use in speech recognition.

Most importantly, LUs in a frame tend overwhelmingly to follow the same annotation structure: the arguments of a lexical unit in a frame consist of the same FEs, regardless of the LU.\footnote{Two or more FEs might be part of a \textit{core set}, yet only one will be realized for valid annotation. For a given frame, different LUs may not realize all the same FEs in all examples.} 
Note the parallel annotation for the two sentences in Figure~\ref{fig:illustration} for the verbs, \textit{buy} and \textit{purchase}.
We use this critical insight in our work, hypothesizing that existing annotation would inform the automated generation of annotation for \textbf{un}annotated LUs in a given frame.


To this end,
\citet{FSautoanno19} defined \textbf{Sister Lexical Unit} as any LU for which annotation exists in a frame, and whose annotation can serve as a model to generate new annotation. A \textbf{Sister} is the LU with annotation and an \textbf{EmptyLU} lacks annotation; the Sister LU and the Empty LU must be of the same part of speech.


To generate augmented labeled data, we first identified all EmptyLUs. For each such LU, we identified a corresponding Sister, specifically, the LU with the greatest number of annotation sets and of the same POS as the EmptyLU in a given frame.
We replaced each occurrence of a Sister in an annotated sentence with the EmptyLU. The replacement process included steps to ensure that the newly generated instance included the correct word forms, i.e., singular and plural forms of nouns or conjugated verbs of a Sister LU, following the strategy that \citet{rastogdurm14} employed, which augmented FN using a paraphrase database \citep{ganitkevitch13ppdb} to add new lemmas and then rewrote existing annotated sentences with those lemmas.


We transferred each sentence in the annotation set of the Sister into that of each EmptyLU, and replaced instances of the Sister with those of the EmptyLU. 
The replacement process ensured that newly generated data included correct word forms, i.e., singular and plural forms of nouns, conjugated verbs, or those with tense marking. 

Ensuring agreement of tense, person, and number between instances of a sister and an EmptyLU required replacing the original word forms with those of the same grammatical number for nouns and tense, person, and number for verbs. 
However, this does not always yield correct results, e.g., for lemmas with irregular plural or past tense forms, as in \textit{ox, oxen} and \textit{bring, brought}.

\subsection{Candidates for Augmentation}
\label{subsec:augmented_data}

The aforementioned process identified 2,805 previously unannotated LUs, of which approximately 500 either were multiword expressions (MWEs) or had an MWE as a potential Sister. 
Thus, the process added at least one sentence to each previously empty annotation set, yielding the potential to provide new annotation for (approximately)
45\% of the unannotated LUs in FrameNet 1.7.
See Table~\ref{tab:examples} in the Appendix~\ref{sec:app:qualitative} for a few pairs of sister and empty LU pairs.
The replacement process did not always yield grammatically correct results. Sometimes FrameNet (arbitrarily) treats MWEs as single LUs, and using a MWE (as a Sister or an EmptyLU) often resulted in ungrammatical sentences. Thus, we also eliminated MWEs from the data in this work, resulting in 2,300 previously unannotated LUs, eligible for augmentation.

\section{Experiments}
\label{sec:experiments}

\begin{table*}[ht]
\small
\centering

\begin{tabular}{lrrrrrrrrr}
\toprule
 & & & \multicolumn{2}{c}{Dev} & \multicolumn{4}{c}{Test}\\
 \cmidrule(lr){4-5}\cmidrule(lr){6-9}
 & \multicolumn{2}{c}{Training Data} & \multicolumn{1}{c}{ArgID} & \multicolumn{1}{c}{FrameID} &   \multicolumn{3}{c}{ArgID} & \multicolumn{1}{c}{FrameID} \\ 
 \cmidrule(lr){2-3} \cmidrule(lr){4-4}\cmidrule(lr){5-5}\cmidrule(lr){6-8}\cmidrule(lr){9-9}
& \#LUs  & \#Lexicographic Annotations & F1 & F1 & P & R & F1 & F1 \\ \midrule
\multirow{2}{*}{\sesame} & 6905 & 164372 & 59.3 & 84.7 & 61.8 & 56.9 & 59.2 & 81.1 \\ 
& 7944 & 260292 & \textbf{61.9} & \textbf{85.3} & \textbf{63.6} & \textbf{60.3} & \textbf{61.9} & \textbf{82.7} \\
\bottomrule
\end{tabular}
\caption{Results comparing the \sesame models trained on the original FrameNet 1.7 data for frame identification (FrameID) and argument identification (ArgID) tasks, with \sesame models trained on our augmented data. 
Data augmentation improves performance on both tasks across all metrics; boldface indicates best performance.
Note that under both settings we consider full-text annotations, a subset of the lexicographic annotations.
Gold-standard frames were used for ArgID, and gold-standard targets for FrameID, in both the baseline and our models.
Development set results only include F1 scores; (by default) \sesame just reports that score.}
\label{tab: Results}
\end{table*} 


To test the idea of leveraging frame-specific annotation to generate new annotation for LUs lacking any, we wish to compare the performance of our approach with a baseline model.
This facilitates determining the extent to which our augmentation algorithm would improve SRL performance, and thus (potentially) contribute to the FrameNet corpus.
Our augmentation approach is applied to FrameNet version 1.7.

\subsection{Training Data}
\label{subsec:data}

In principle, our data augmentation algorithm (\S\ref{sec:methodology}) could provide new annotation for 2,300 LUs (\S\ref{subsec:augmented_data}), all of which could be used for training.
However, we wish to determine whether using augmented data improves performance on the existing test sets. 
Additionally, care must be taken to avoid the logical flaw of using augmentation algorithm from producing labeled sentences specifically for the test set: the process would circularly assume that the algorithm offers accurate labelling. 

We addressed this issue by adjusting the training data for the baseline: we removed annotation from 1,500 randomly selected LUs from the lexicographic portion of the training data for FN 1.7.
Each of these 1,500 LUs functions as EmptyLUs when stripped of their annotation. 
Our baseline model was trained with the remaining lexicographic annotations. 
Our full model was trained on data given by the augmentation algorithm to re-annotate the 1,500 LUs, in addition to the data used for the baseline.
Note that not all LUs are eligible for training in \sesame, which involves a pre-processing step to remove those LUs incompatible with the parser; Table~\ref{tab: Results} provides the final training data statistics.

In addition to the lexicographic annotations, both the baseline and the augmented setup use all the full-text annotations in FrameNet 1.7.
Our training data adjustments above allow using the standard validation and test sets for FrameNet 1.7, following \citet{DBLP:journals/corr/SwayamdiptaTDS17}.


\subsection{Model and Hyperparameters}
\label{subsec:model}

All of our experiments were performed using the non-syntactic model provided in \sesame \citep{DBLP:journals/corr/SwayamdiptaTDS17, swayamdiptaEtAl18syntactic}.
We follow the basic setup in \sesame which assumes gold targets for FrameID and gold frames for ArgID.
We left most hyperparameters associated with \sesame at their default values, but decreased the learning rate for training the ArgID stage to $0.0001$ from its original value of $0.0005$. 
The decrease allowed training ArgID on the entire dataset in three epochs without error, which serves reproducibility requirements.

\subsection{Results}
\label{sec:results}

Table \ref{tab: Results} presents test and validation set performances comparing two \sesame models, trained with and without the augmented data for ArgID and FrameID. 
All measures in the experimental setting outperformed the baseline, highlighting the value of data augmentation. 
Although our main interest was testing the system's performance for ArgID (i.e., identifying semantic roles / FEs for given frame-LU annotations) of the augmented data, we also observed improvement in the identification of frames.
The results in the experimental setting show an absolute improvement of 2.7 in F1 on ArgID and 1.6 in F1 on FrameID.\footnote{We further experimented with cross validation settings for both FrameID and ArgID; however, the runtime of \sesame was prohibitively slow in this setting.}

\subsection{Error Analysis and Discussion}
\label{analysis}


At times the augmentation algorithm produced ungrammatical sentences. 
We do not believe that errors occurred because of ungrammatical sentences. Still, FrameNet would not conjure producing data with ungrammatical sentences.\footnote{A complete manual analysis of the data remains to be done. In future extensions of this work, we will report such an analysis, including the percentage of grammatical sentences.}
A manual analysis of a random sample of LUs showed three categories of errors: (a) word form mismatch; (b) incorrect or missing marker; and (c) semantic mismatch, examples of which appear in (a) - (c):

\begin{description}
  \setlength{\itemsep}{1pt}
  \setlength{\parskip}{0pt}
  \setlength{\parsep}{0pt}
  \item (a) *The moon was now \textit{occlude} by clouds.\label{wf}
  \item (b) *And he \textit{complained} the endless squeeze on cash.
  \item (c) ?The faint \textit{flash} from a street light showed him the outline of a hedge.\label{sem}
\end{description}

\noindent
The semantic mismatch in (c) holds between \textit{faint}.a and \textit{flash}.n, yielding a semantically odd sentence not an ungrammatical one. 
By definition, a \textit{flash} would not be \textit{faint}.\footnote{The results showed several types of semantic mismatch.
Space limitations preclude discussing all of them here.}

Recall that incorrectly identifying a frame necessarily yields incorrect FEs (ArgID). The analysis showed that errors occurred with highly polysemous lemmas, such as \textit{make}.v, where FrameNet has 10 LUs; in (d), \sesame misidentified 
\textit{making}.v as \texttt{Manufacturing}, not \texttt{Earnings\_and\_losses}. Similarly, (e) is a misidentification of \textit{gathering} as \texttt{Food\_gathering}, not \texttt{Come\_together}. Note the metaphorical use of the verb in (e), compared to literal senses of the lemma \textit{gather}.v. 

\begin{description}
  \setlength{\itemsep}{1pt}
  \setlength{\parskip}{0pt}
  \setlength{\parsep}{0pt}
  \item (d) I like working and \textit{making} money.
  \item (e) ...the debate on welfare reform is \textit{gathering} like a storm at sea...
\end{description}

\section{Related Work}
\label{sec:related}

Data augmentation in the context of FrameNet and SRL is not novel. \citet{pavlick15FN} created an expanded FN via automatic paraphrasing and crowdsourcing to confirm frame assignment of LUs.
\citet{hartmannassess17} used automatically generated training data
by linking FrameNet, PropBank, and VerbNet to study the differences among those resources in frame assignment and
semantic role classification. The results showed that the augmented data performed as well as the training data for these tasks on German texts.
With similar goals to those of this work, \citet{AugmentFN20} proposed augmenting FN annotation using notions of lexical, syntactic, or semantic equivalence, and FN's frame-to-frame relations (Inheritance, Using, SubFrame, etc.) to \textit{infer} annotations across related frames, producing a 13\% increase in annotation. 

\section{Conclusion and Future Work}
\label{sec:future}

We outlined an approach to Frame-SRL that seeks to leverage existing annotation in a frame to generate new annotation for previously unannotated LUs in the same frame, by exploiting the already identified target and frame of a given lemma.
Further, we demonstrated that augmentation of data improves Frame-SRL performance substantially.
While we show the advantages of this approach on the open-SESAME baseline, future work might involve studying if using this type of data augmentation will yield improvements on a state-of-the-art SRL system, e.g. \citet{FeiEtAl21}.

Several other interesting directions for future work exist, including the following: refining the augmentation algorithm to produce more grammatical sentences; streamlining open-SESAME's code to run more efficiently; use active-learning to inform the data generation and to assist in the SRL task; and experimentation with other computational techniques, as suggested by \citet{FSautoanno19}.
These plans will advance defining processes to include Frame-SRL into FrameNet's development process, the ultimate goal of this work.\footnote{Rigorous testing of multiple factors will be required to evaluate the viability of including Frame-SRL into FrameNet's existing primarily manual process.}

\section*{Acknowledgements}
The authors are grateful to the three anonymous reviewers for their feedback, and to Nathan Schneider for his input on various aspects of the experiment, including very early conversations about pursuing the idea. Collin Baker and Michael Ellsworth have contributed to the completion of the paper in countless ways.

\bibliography{anthology,custom}
\bibliographystyle{acl_natbib}

\appendix
\label{sec:appendix}

\section{Qualitative Examples}
\label{sec:app:qualitative}

Table~\ref{tab:examples} provides lexicographic annotations with sister LUs and their corresponding empty LUs, the latter being populated by our approach.

\begin{table}[ht]
\centering
\small
   \begin{tabular}{p{3.15cm}cp{3.15cm}} 
        \toprule 
        \bf Sister Lexical Units & &\bf Empty Lexical Units \\ \midrule
\textcolor{BrickRed}{He}\textcolor{BrickRed}{$_{\textsc{Agent}}$} 
 {\textit{stamped}}  \textcolor{Green}{his foot}   \textcolor{Green}{$_{\textsc{Body\_part}}$} \textcolor{blue}{into his flying-boot} \textcolor{blue}{$_{\textsc{Peripheral}}$} . & \multirow{3}{*}{$\rightarrow$} &
  \textcolor{BrickRed}{He}\textcolor{BrickRed}{$_{\textsc{Agent}}$} 
 \textit{\textcolor{black}{bended}}
 \textcolor{Green}{his foot} 
 \textcolor{Green}{$_{\textsc{Body\_part}}$}
 \textcolor{blue}{into his flying-boot} \textcolor{blue}{$_{\textsc{Peripheral}}$} . \\
 \midrule[0.03em]
John-William , who knew that he would have been a Chartist himself had he remained a \textit{\textcolor{black}{poor}} \textcolor{blue}{man} \textcolor{blue}{$_{\textsc{Person}}$} , felt sorry about that death . &  \multirow{5}{*}{$\rightarrow$} &
 John-William , who knew that he would have been a Chartist himself had he remained a
 \textit{\textcolor{black}{rich}}
 \textcolor{blue}{man} \textcolor{blue}{$_{\textsc{Person}}$} 
  , felt sorry about that death .\\
   \midrule[0.03em]
 Because of the censorship , and the obvious need to avoid 
  \textcolor{RedOrange}{dangerously} \textcolor{RedOrange}{$_{\textsc{Degree}}$}
  \textit{\textcolor{black}{critical}}
  \textcolor{DarkOrchid}{comments} \textcolor{DarkOrchid}{$_{\textsc{Expressor}}$}
  \textcolor{blue}{about the regime and the war} \textcolor{blue}{$_{\textsc{Evaluee}}$}
  , the correspondence to and from the front provides no easy guide to political attitudes . &  \multirow{10}{*}{$\rightarrow$} &
   Because of the censorship , and the obvious need to avoid 
  \textcolor{RedOrange}{dangerously} \textcolor{RedOrange}{$_{\textsc{Degree}}$}
  \textit{\textcolor{black}{commendable}}
  \textcolor{DarkOrchid}{comments} \textcolor{DarkOrchid}{$_{\textsc{Expressor}}$}
  \textcolor{blue}{about the regime and the war} \textcolor{blue}{$_{\textsc{Evaluee}}$}
  , the correspondence to and from the front provides no easy guide to political attitudes . \\ 
   \midrule[0.03em]
 When I did eventually tell her
  \textcolor{violet}{she} \textcolor{violet}{$_{\textsc{Experiencer}}$}
  was
  \textcolor{RedOrange}{really} \textcolor{RedOrange}{$_{\textsc{Degree}}$}
  \textit{\textcolor{black}{embarrassed}}
  , and tried telling me that I was making it up ! &\multirow{5}{*}{$\rightarrow$} &
  When I did eventually tell her
  \textcolor{violet}{she} \textcolor{violet}{$_{\textsc{Experiencer}}$}
  was
  \textcolor{RedOrange}{really} \textcolor{RedOrange}{$_{\textsc{Degree}}$}
  \textit{\textcolor{black}{tormented}}
  , and tried telling me that I was making it up !  \\
   \midrule[0.03em]
  This regulation prevented 
 \textcolor{blue}{US banks} \textcolor{blue}{$_{\textsc{Theme}}$}
\textit{\textcolor{black}{located}}
\textcolor{Green}{in the US} \textcolor{Green}{$_{\textsc{Location}}$}
, but not abroad, from paying interest on deposits above a given rate . & \multirow{6}{*}{$\rightarrow$} &
 This regulation prevented 
 \textcolor{blue}{US banks} \textcolor{blue}{$_{\textsc{Theme}}$}
\textit{\textcolor{black}{situated}}
\textcolor{Green}{in the US} \textcolor{Green}{$_{\textsc{Location}}$}
, but not abroad, from paying interest on deposits above a given rate . \\
  \bottomrule
\end{tabular}
\caption{
Parallel lexicographic annotations of sister LUs (expert-annotated) and emtpy LUs (augmented by our approach). 
Frame Element (FE) names are in \textsc{Small Caps} and their instantiations are color-coordinated; and LUs or targets are in \textit{italics}.
}
\label{tab:examples}
\end{table}

\end{document}